# AN ONTOLOGY-DRIVEN FRAMEWORK FOR SUPPORTING COMPLEX DECISION PROCESS

## JUNYI CHAI and JAMES N.K. LIU

*Department of Computing,*
*The Hong Kong Polytechnic University*
*Hung Hom, Kowloon., Hong Kong, SAR*
*{csjchai,csnkliu}@comp.polyu.edu.hk*

ABSTRACT - The study proposes a framework of ONTOlogy-based Group Decision Support System (ONTOGDSS) for decision process which exhibits the complex structure of decision-problem and decision-group. It is capable of reducing the complexity of problem structure and group relations. The system allows decision makers to participate in group decision-making through the web environment, via the ontology relation. It facilitates the management of decision process as a whole, from criteria generation, alternative evaluation, and opinion interaction to decision aggregation. The embedded ontology structure in ONTOGDSS provides the important formal description features to facilitate decision analysis and verification. It examines the software architecture, the selection methods, the decision path, etc. Finally, the ontology application of this system is illustrated with specific real case to demonstrate its potentials towards decision-making development.

## 1. INTRODUCTION

Decision Support Systems (DSS) have been proposed since the late 1960s to help decision maker improve the efficiency and correctness in decision making. Along the development of DSS, researchers notice that the decision making in reality is not just individual decision but often involving multiple peoples. As a matter of fact, many decision problems (such as great strategic decision of government or industry, the managing decision of large company), have the complex internal structure and in need of making decision by a large decision group with complex relationship among people. To address these problems, we provide a group decision process structure and system framework coupled with relatively complex decision groups and tasks.

In section 2, we investigate the task decomposition and group selection process based on the analysis of the ontology structure in the application domain, in order to reduce the complexity of them. Section 3 provides the design of a sub-system named Workshop System, in order to resolve the conflict in group decision process. It describes the use of ontology approach and metasynthesis methodology [10] for designing the group argumentation models. Section 4 leads to the ontology-driven system framework including the overall group decision process, system architecture, and ONTOGDSS hierarchical structure with ontology-based decision resource layer. Finally, section 5 presents an ontology application in decision-problem domain with illustrative examples.

## 2. DECISION PROCESS

### *2.1 Task Decomposition*

Ontology is defined as "a set of knowledge terms, including the vocabulary, the semantic interconnections and some simple rules of inference and logic, for some particular topic" [1]. That is to say, ontology captures the model of knowledge for a particular domain. They allow us to describe resources on the web and the relationships between those resources. Accordingly, ontology can be regarded as metadata which play an important role in decision process. System provides the methods for generating a series of alternatives for comparison and evaluation of different decision-makers. Thus, ONTOGDSS relies on metadata to describe the attributes, objectives, context, constraints, types, criteria of the complex decision problem in real world, and therefore will be ontology-driven. So, it is necessary to develop ontologies which can encode the semantic representation of the structural complex decision problem, in order to form a specific, clear decision path.

Based on Herhert A Simon's [2] dichotomy of decision problem, we develop the idea of dividing decision problems into three categories: structural problem, semi-structural problem and non-structural problem. For structural problems, we can load decision models, methods, data and other information as reference. For other two problems, since semi-structural and non-structural problems mean that they have never been shown up before and usually presented as qualitative textual form/document with complex semantic structure, therefore, besides loading necessary data in database, it is important to make the reference via ontology-approached knowledge management system in various decision domains.

The ONTOGDSS is designed as an ontology-based intelligent information system platform. It highlights the needs for considering contextual aspects in system perspective. Besides, ontology in specific decision-problem domains would include basic concepts such as decision targets, principles, limitations, and additional concepts of problem style, characteristics, evaluation criteria and etc. Therefore, problem representative and description in ontology approach are not only important to those structure-problems for better searching and matching in previous models or methods, but also used especially for those semi-structure/non-structure problems for group decision process.

*2.2 Group Selection*

DSS ontology can be defined as formal descriptions of decision concepts by basic terms and relationships as well as the rules for combining these terms in a certain problem domain. While abstraction of an ontology development is similar to definition of a conceptual model, the focus is on extended definitions of relationships and concepts, and having the explicit goal of reuse and sharing knowledge by using a common framework. In GDSSs, the concept of decision-group usually is presented in contextual form with complicated relationship and structure. However, the concept of group is usually defined in literature as a kind of individual-aggregated entity which does not depend on individual properties with conceptualization. This paper analyzes and establishes the decision-group through ontology-based conceptual extraction in contextual decision-group domain. This approach can eliminate the confusions associated with the term "Group". Once various structures are established, the unique characteristics of each would be emerged. Thus, researches can be focused on the various interactions among participants as well.

Based on literature review of "Group" concept, and previous group selection methods [3], we provide a Double Selection Model to process group selection. It requires decision group to be selected in two aspects at least. For example, we need to evaluate the work performance of four peoples (alternatives) $Y_i = \{Y_1, Y_2, Y_3, Y_4\}$ ($i = 1,2,3,4$) by five suitable evaluators (decision maker) $d_{ij}^*$, who are respectively from five different parts: higher authorities $G_1$; peer authorities $G_2$; lower authorities $G_3$; independent people outside of the company $G_4$; alternatives themselves $G_5$; where $G_j = \{G_1, G_2, G_3, G_4, G_5\}$ ($j = 1,2,3,4,5$). For alternative $Y_i$, the suitable evaluators $d_{ij}^*$ are respectively selected from five different parts $G_j$ through the Double Selection Model. First selection can base on the decision task types, and second selection can base on decision maker's characters. After these processes, evaluator candidates $d_{ij}$ ($d_{ij} \in D_j$) of alternative $Y_i$ can be selected to be the suitable evaluators $d_{ij}^*$. The key issue of this approach is to establish a proper assessment criteria system. Once it is established, many classic multi-criteria decision-making approaches can be adopted to solve this problem, such as outranking relations approaches including ELECTRE [4] and PROMETHEE [5], or preference disaggregation approaches including UTA [6]. In this example, the criteria of first selection can be set as the different professional fields: computing, economic, management. And the other one can be set on individual characteristics of decision makers: age, sex, nationality, education background, etc.

## 3. WORKSHOP SYSTEM

*3.1 Argumentation based on Ontology Approach and Metasynthesis Methodology*

Argumentation has become a keyword of Artificial Intelligence, especially in sub-fields such as multiple-source information system with natural language processing. One of the abstract frameworks of Argumentation system is Dung's one [7] which shows that several formalisms for non-monotonic reasoning can be expressed in terms of this argumentation system. Ontology technique can be used to model natural language for data integration, data interoperability and data visualization. By using this, humans and computers (software agents) can have a consensus on the resource structure [8]. In the past, ontology approaches have been a universal technique to build explicit understanding of the structure of complex problem such as those in World Wide Web design, medical informatics, bioinformatics and geospatial informatics [9]. In these cases, ontology was not only used for data integration and interoperability, but also for outlining system metadata. In this study, based on semantic ontology, we try to establish a workshop system framework for argumentation processes.

Workshop system is, for specific complicated problem, a kind of Meta-synthetic process from qualitative to quantitative, which integrated the knowledge and intelligence of expert group, data, and useful equipments. In this paper, argumentation process orients to the complex decision-problem and group structure. Therefore, it is necessary to apply the metasynthesis methodology to design the workshop system for more efficient decision processes. In the design of decision processes, experts establish some qualitative and non-precise thinking or ideas based on the availability of synthetic knowledge. Through ontology representation process, such information can be clearly described or defined, and form the quantitative expression. By this express process from qualitative to quantitative, most of the knowledge which is used in group decision process can be rationally represented and verified. In fact, the problem-solve process is also from qualitative to quantitative. Therefore, this qualitative knowledge, useful information or other knowledge in expert's mind are raised to the quantitative reorganization as whole by organization, synthesis, model establishment, iterative evaluation and modification.

*3.2 Group Argumentation Model*

In workshop system, the participants in argumentation process are constructed as a group. From the view of Metasynthesis methodology [10], the integration of human's qualitative intelligence and computer's quantitative intelligence is one feasible processing method to solve complex problem in reality. We notice that, previous argumentation models did not include the properties of decision task and did not consider the particularities of complex decision task. However, in reality, these factors are very important for complex-task oriented decision making. Therefore, the paper proposes a multi-layer structural group argumentation model as shown in Figure I.

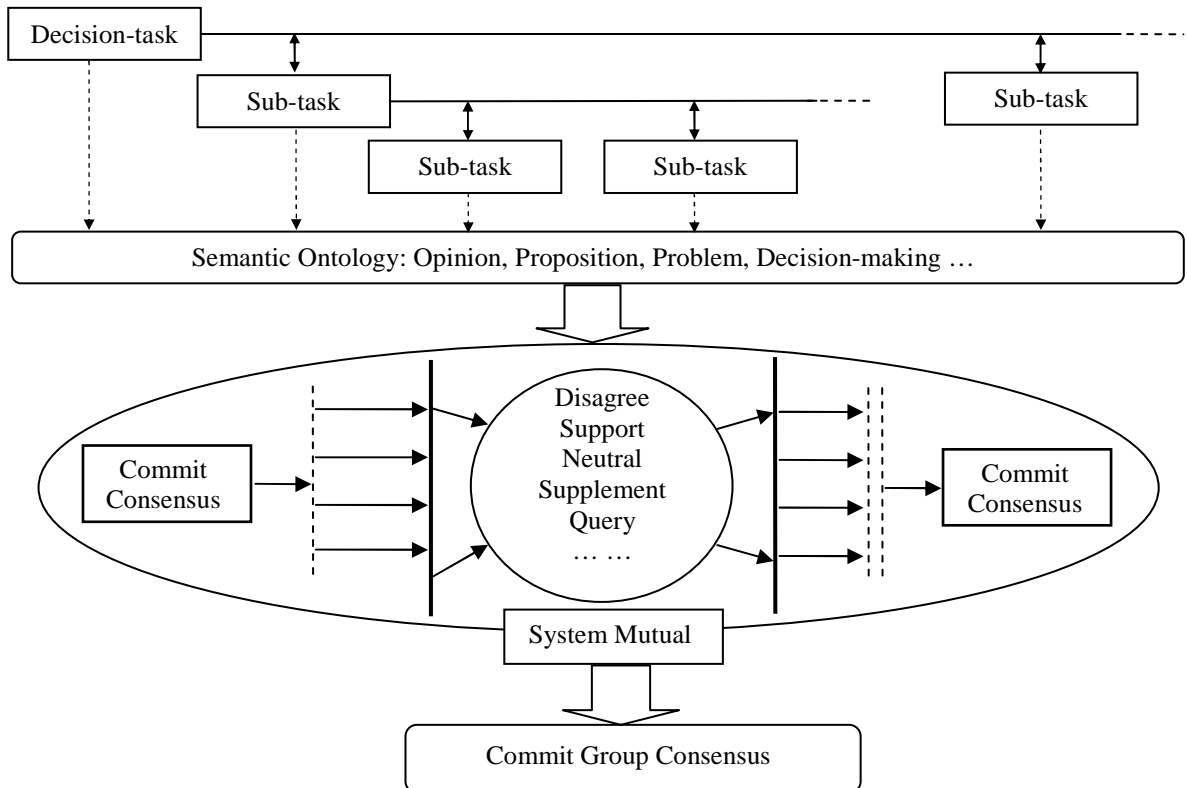

**Figure I. Multi-layer structural group argumentation model.**

Through ontology analysis, system extracts semantic objects as opinions, proposition, problems and etc, to form basic elements for argumentation. Then, these basic information elements input into Workshop system and interact with others. In this group argumentation model, we define five kinds of basic relations between information elements to model the interactions. They are Disagree, Support, Neutral, Supplement, and Query. Finally, system commits the consensus and feedback these results for following decision making section.

## 4. ONTOLOGY-DRIVEN FRAMEWORK FOR GROUP DECISION PROCESS

*4.1 Group Decision Process*

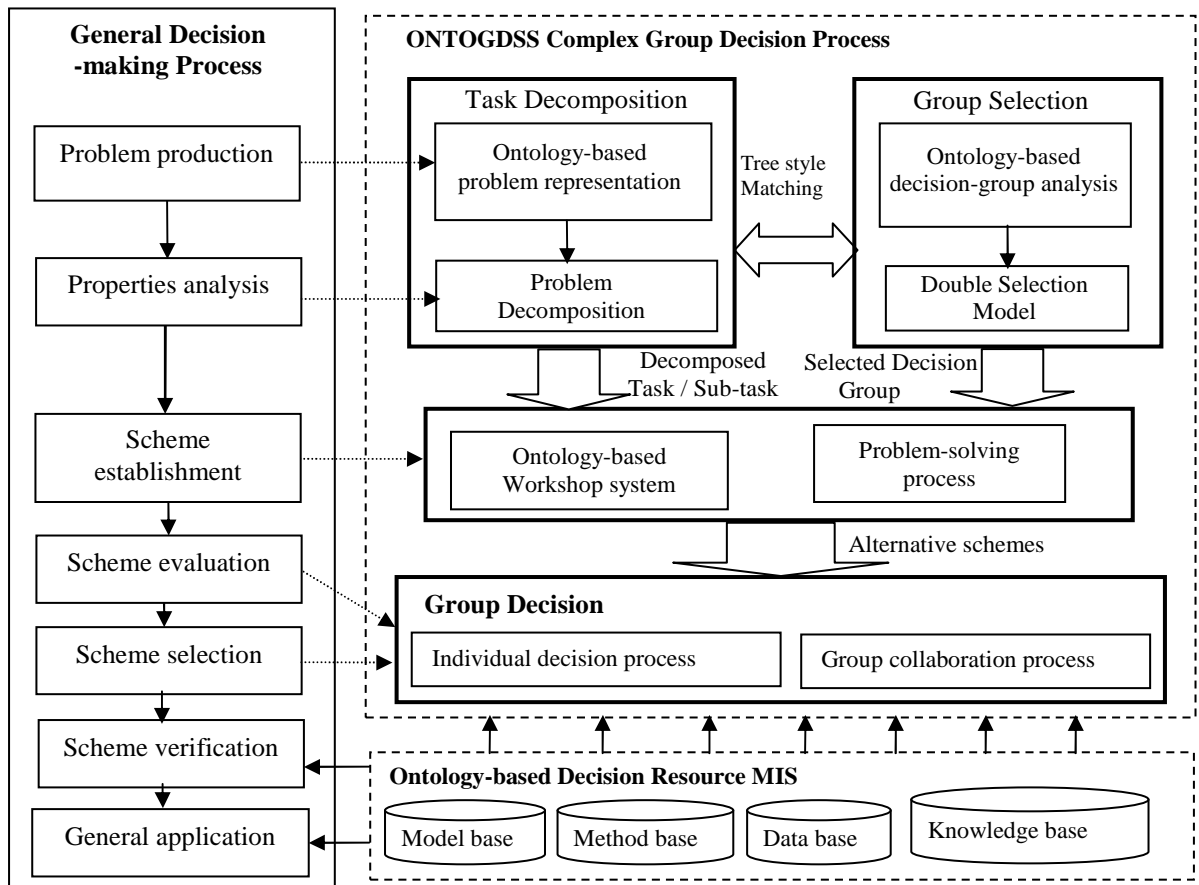

**Figure II. Ontology-driven complex group decision process.**

In this section, the ONTOlogy-based Group Decision Support System (ONTOGDSS) is presented. This system framework consists of two aspects: group decision process, system hierarchical structure. All of them are based on ontology driven representative and description of decision-problem.

Figure II shows the complex group decision process in ONTOGDSS. At first, we summarize the general process of decision-making as seven stages including (1) problem production (2) properties analysis (3) scheme establishment (4) Scheme evaluation (5) Scheme Selection (6) Scheme verification (7) General application. Following this general process, our proposed decision process considers two important situations. First, this process is used to figure out the complex decision task. Second, it is used for the complex large decision group. Oriented by these two situations, we design the Group argumentation process and problem-solving process to establish alternative schemes. And through group decision algorithm, the system selected the alternative schemes. Besides, the ontology-based decision resource MIS provide the support in data accessing and information storage.

*4.2 System Hierarchical Structure*

For processing the complex group decision, the structure of ONTOGDSS includes four layers.
1. Task decomposition layer

Based on ontology-approached representation and description of decision-problem, we can clarify its properties and limitations. Then, a tree-like decision-task structure is formed after confirming the decomposition direction. In general, task decomposition process in this layer provides the important basis and targets, and also provides some alternative decision paths.

2. Decision problem-solving layer

The system needs to organize all useful experts (or selected decision-people) to solve the task and finally form a set of problem-solving schemes, and storage into the corresponding scheme base. In this process, the workshop system which is a sub-system of ONTOGDSS and with a useful problem-solving method provides systematical supports to ontology-based group argumentation process.

3. Group decision layer

This layer includes individual decision process and Group collaboration process. The main responsibility of this layer is to appoint task via mathematical algorithms, allow decision-makers to rank alternative schemes and commit a consensus at last. Through the summarization of the whole decision process and final results, we can obtain the most satisfactory scheme for this appointed task/node.

4. Ontology based Decision-resource layer

Note that this layer is based on ontology approach. As we mentioned above, for structural decision-problem, Model MIS and Method MIS can provide the model and method of previous decision experience, case or theory. Based on ontology-approach such as semantic extraction [11], we can represent and describe these decision problems which are stored in corresponding bases.

## 5. APPLICATION

This application focuses on ontology-based decision problem representation and extraction. In reality, decision problems are complex, uncertain, and dynamic. Ontologies as metadata provide the approach to represent and construct decision problem, so that its structure, characteristics, properties and limitations can be analyzed accordingly. In this paper, we apply the ontology-based brokering service Ontobroker [12] to construct the representation of decision problem as shown in Figure III.

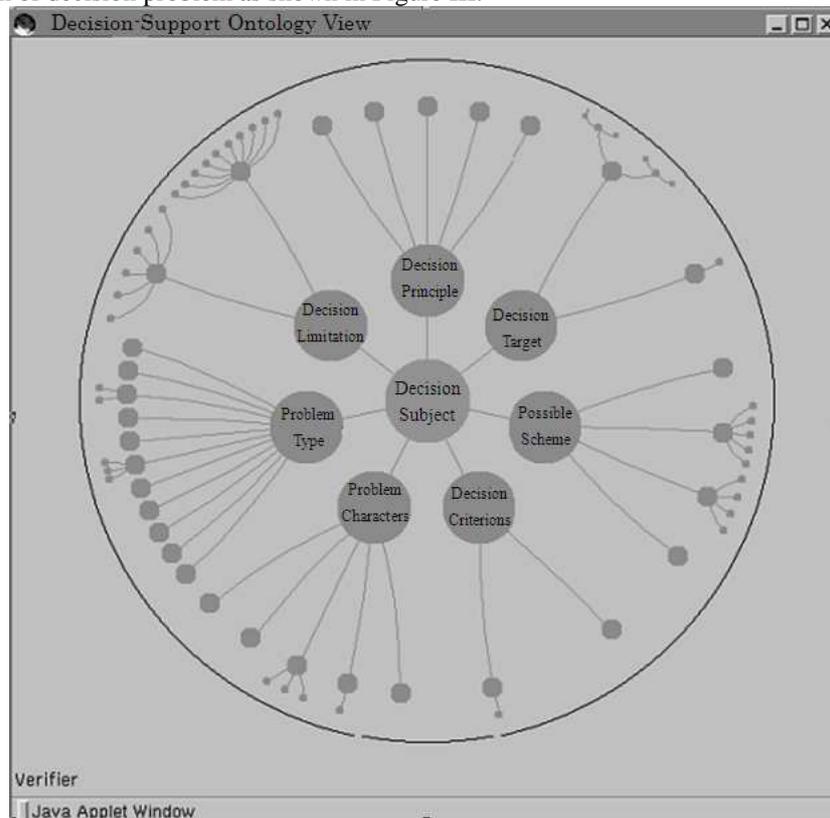

**Figure III. Ontology based decision-problem construction.**

**Figure IV. Ontology query interface of one node.**

From Figure III, decision subject is extracted into seven concepts: Problem Type, Decision Limitation, Decision Principle, Decision Target, etc. As Bui and Bodart [13] mentioned, one decision problem can be treated as an action for achieving some target, and considering the characteristic elements of every decision problem in two factors: (1) basic factors such as limitations, principles, targets; (2) additional factors such as decision type, characteristics, criterions, and scheme. Therefore, we extract these seven concepts from decision subject for further analysis. Figure IV shows the ontology query interface of one node. Particularly, in "class" column, we classify these extracted concepts into "Basic Factor" and "Additional Factor" as mentioned in above.

## 6. CONCLUSION

This paper proposes an ontology-driven complex group decision process and corresponding decision support system named ONTOGDSS. We firstly present an ontology-based construction approach according to the complex structure of group and task in reality. Then, based on ontological problem representation, we provide designs of group decision process and the framework of support system. Finally, we present an implementation on ontology-based decision problem representation and extraction. In future work, we will make effort to develop more intelligent middleware and groupware of ONTOGDSS. We also will consider how to adopt ontology-driven knowledge/data mining technologies to develop decision-resource MIS on our proposed framework. Besides, how to extend ONTOGDSS for solving uncertainty group decision making problems would be a possible direction.

**Acknowledgement**

The authors would like to acknowledge the partial supports from the GRF 5237/08E of the Hong Kong Polytechnic University.